\documentclass[letterpaper, 10pt, conference]{ieeeconf}      %

\IEEEoverridecommandlockouts                              %

\usepackage{amsmath} %

\title{\LARGE \bf
	Investigating Outdoor Recognition Performance of Infrared Beacons for Infrastructure-based Localization
}

\author{Alexandru Kampmann, Michael Lamberti, Nikola Petrovic, Stefan Kowalewski and Bassam Alrifaee%
	\thanks{Alexandru Kampmann, Michael Lamberti, Nikola Petrovic, Stefan Kowalewski and Bassam Alrifaee are with the  Chair for Embedded Software, RWTH Aachen University, 52074 Aachen, Germany
		{\tt\small \{kampmann, lamberti, petrovic, kowalewski, alrifaee\}@embedded.rwth-aachen.de.}}%
}

\usepackage[utf8]{inputenc}
\usepackage{mathtools}

\DeclareMathOperator{\sgn}{sgn}
\usepackage{todonotes}
\usepackage{booktabs}
\usepackage{cite}
\usepackage{hyperref}
\usepackage[binary-units=true]{siunitx}
\DeclareSIUnit\pixel{px}

\begin{document}

	\maketitle
	\thispagestyle{empty}
	\pagestyle{empty}

	\begin{abstract}
		This paper demonstrates a system comprised of infrared beacons and a camera equipped with an optical band-pass filter.
		Our system can reliably detect and identify individual beacons at \SI{100}{\m} distance regardless of lighting conditions.
		We describe the camera and beacon design as well as the image processing pipeline in detail.
		In our experiments, we investigate and demonstrate the ability of the system to recognize our beacons in both daytime and nighttime conditions.
		High precision localization is a key enabler for automated vehicles but remains unsolved, despite strong recent improvements.
		Our low-cost, infrastructure-based approach is a potential step towards solving the localization problem.
		All datasets are made available here \url{https://embedded.rwth-aachen.de/doku.php?id=forschung:mobility:infralocalization:concept}.
	\end{abstract}
	\section{INTRODUCTION}
	\textbf{Motivation:} High-precision localization is a key enabler for automated vehicles, as most disclosed automated vehicle (AV) prototypes rely on a high-precision map of the environment they operate in.
	These maps contain static information about the environment (e.g. lane boundaries, traffic light locations, stop sign, traffic rules) and are generated before driving autonomously in a specific area.
	These pieces of information are often required ahead of time for trajectory planning and are difficult to extract from live sensor data with sufficient foresight due to sensor range, occlusions or non-line-of-sight.
	Localization within these maps typically is either based on heuristic features that have been stored during map generation or based on Global Navigation Satellite System (GNSS) approaches, often in combination.
	Currently, relevant static environment elements are retrieved from the map and combined with dynamic elements of the environment extracted from live sensor data, e.g., other traffic participants, free space, state of traffic lights.
	This environment model is the basis for subsequent decision making, trajectory planning and control.
	\\
	A variety of localization approaches for autonomous vehicles have been proposed and used in practice.
	The majority of the proposed approaches are following two fundamental ideas: GNSS-based systems, and systems based on heuristic features.
	\begin{figure}[t]
		\centering
		\includegraphics[width=\linewidth]{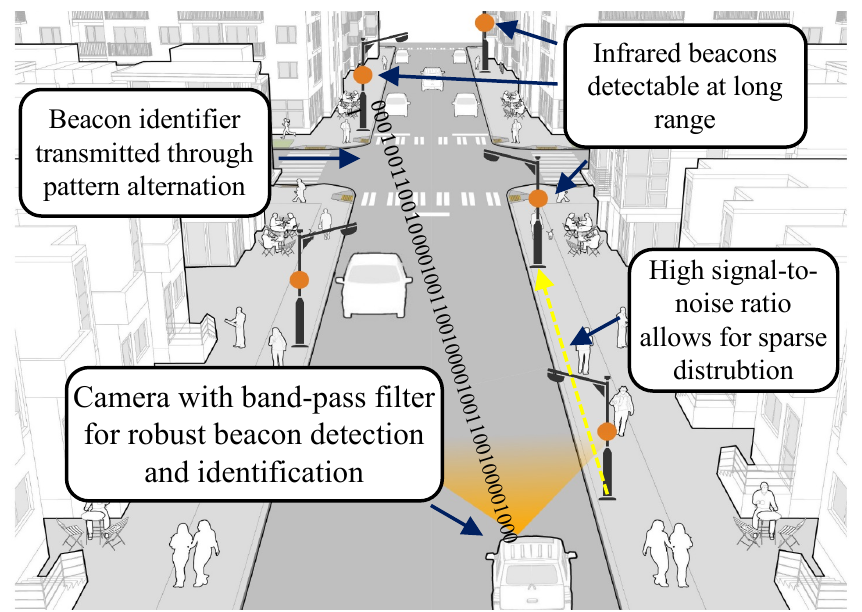}
		\caption{Our proposed system uses sparsely distributed infrared beacons that transmit an identifier allowing the vehicle to localize itself. The detection distance is multiples longer than the typical distance between light poles in urban areas in Germany. }
		\label{fig:concept}
	\end{figure}
	GNSS-based systems, such as GPS or Galileo, use satellite signals, to which direct line of sight is required, to triangulate their position.
	The requirement for line of sight poses a problem in tunnels or urban environments, where tall buildings may reflect or block the signal, culminating in reduced accuracy or inability to localize. 
	These issues can be somewhat mitigated by fusing the GNSS position with measurements from other sensors, such as an inertial measurement unit (IMU) and map data.
	Such approaches can help to temporarily overcome situations with degraded performance, but suffer from IMU drift and do not reliably solve the robustness problem in general.
	As detailed in Section \ref{sec:related_work}, camera or LIDAR-based localization often relies on finding and identifying heuristic feature points in images.
	However, these approaches suffer from weak long-term feature stability in different weather and lighting conditions.\\
	\textbf{ Contribution:} As illustrated in Fig. \ref{fig:concept}, we investigate the idea of augmenting the infrastructure with active visual features that are specifically designed to be easily detectable and identifiable at long-range.
	As detailed in Section \ref{sec:related_work}, approaches proposed in prior-art are not detectable at long ranges and therefore require a high density, rendering them impracticable for large-scale outdoor environments.
	Once the beacon has been detected and identified, subsequent processing steps towards localization are very similar to existing, feature-based localization schemes \cite{rabinovich_cobe_2020}.
	As the performance is heavily influenced by feature matching and recognition, our work primarily investigates the ability of the system to detect, track and identify beacons in outdoor scenarios at long range.
	This work presents proof-of-concept for a camera with band-pass filter that is able to detect and identify single infrared beacons at \SI{100}{\meter} range at both day and night.
	\\
	Due to the large detection distance, sparse distributions of our unobtrusive low-cost beacons is sufficient to augment wide areas for autonomous vehicle deployments or other robotic applications.
	For example, our detection distance is three times greater than the typical distance between light poles in urban areas in Germany.
	Our contributions are summarized as follows:
	\begin{itemize}
		\item We propose a simple and robust approach for infrastructure-based keypoints based on sparsely distributed infrared beacons. 
		\item We describe the hardware and protocol design and image processing pipeline in detail.
		\item We provide a proof-of-concept evaluation for a long-range outdoor scenario for both day and night.
	\end{itemize}
	\section{Related Work}
	\label{sec:related_work}
	\textbf{GNSS approaches} such as GPS or Galileo use satellite signals to determine the current position.
	While GPS achieves meter accuracy\cite{tan_dgps-based_2006-1}, performance can be improved through the use of differential GPS (DGPS) or real-time kinematic (RTK) solutions, which use base-stations with known location to improve positioning accuracy \cite{kuutti_survey_2018}. 
	The reliability problem remains in urban environments or tunnels, where signals can be reflected or blocked entirely, culminating in reduced accuracy or inability to localize.
	To overcome situations with degraded GNSS availability, inertial measurement units (IMUs) have been used to estimate the vehicle position relative to an initial position.
	For example, Zhang et al. \cite{zhang_sensor_2012} improve raw GPS performance through fusion with an IMU, but the accumulated root mean square positioning error after \SI{408}{\meter} is at \SI{7.4}{\meter}, which is too high for automated driving.\\
	\textbf{Camera-based techniques} have been investigated in numerous publications \cite{li_location_2010,sattler_hyperpoints_2015,torii_247_2015,taira_inloc_2018, brachmannVisualCameraReLocalization2021}.
	Often, these techniques rely on heuristic feature points, which are embedded in a map and matched with features extracted from live camera images for localization.
	In \cite{sattler_benchmarking_2018} Sattler et al. benchmark various approaches under changing environment conditions and conclude \textit{'that long-term localization is far from solved'}.
	Problems arise from feature detection and matching with corresponding map features as the scene appearance changes over time.
	Frequent map updates and semantic information may accommodate for long-term changes\cite{Schoenberger2018Semantic,Xiao2018MonocularSemantic}, but performance issues under different lighting and weather situations remain.\\
	\textbf{Infrastructure-based approaches} try to alleviate recognition problems of heuristic features by purposely placing hand-crafted features or devices in the environment.
	We first present related work for indoor systems and then discuss prior work for outdoor applications.\\
	Li et. al. propose Epsilon, an indoor positioning system that uses regular LED lamps \cite{li_epsilon:_2014} and that achieves sub-meter accuracy indoor.
	The authors of \cite{Hijikata2009indoor-led} use infrared LEDs placed on walls, which are detected using a camera system, allowing for mobile robot localization. 
	To simplify detection, an optical low-pass filter is used in front of the camera, which blocks visible light.
	Different to our work, the LEDs do not communicate an identifier and are not distinguishable.
	Fiducial tags such as arUco \cite{garrido-jurado_generation_2015} and AprilTags \cite{olson_apriltag_2011} are purposely designed for robust recognition.
	Due to their passive nature, they are difficult to detect and identify at long-range and require high density.
	In contrast, our we try to minimize the number of beacons required, potentially rendering the idea economically feasible and avoiding environmental clutter.
	
	The Vehicle Information and Communication System (VICS) in Japan consists of over 56.000 infrared transceivers that are used for traffic monitoring and traffic information purposes.
	The beacons are mounted above the road, have a communication range of \SI{3.5}{\meter}.5
	In \cite{hayama_advanced_nodate}, the authors were able to extend the communication range to \SI{6}{\meter}.
	Our proposed system is related to the Visible Light Communication (VLC) domain \cite{yamazato_overview_2017}, for which we will now present related work.
	In \cite{eso_experimental_2019} the authors modulate traffic light intensities and capture and decode the transmitted signal using a camera.
	The authors do not test their system in motion and traffic lights are too sparsely distributed to be used for localization in wider areas.
	The authors of \cite{yamazato_image-sensor-based_2014} also investigate the use of traffic lights as communication channels and achieve transmission rates of multiple \SI{}{\mega\bit\per\second}.
	The same authors design a system using a LED-based transmitter capable of high-speed transmissions, but rely on an expensive high-speed camera and much larger beacons \cite{nagura_improved_2010}, \cite{nagura_tracking_2010}.
	In \cite{Liu2003positioning-beacons}, Liu et. al. propose an outdoor localization system incorporating LEDs in traffic lights and visible light beacons.
	In contrast to our work, the use of optical band-pass filters was not investigated and no outdoor evaluation was conducted.
	Kim et al. propose a VLC system for localization but the demonstration and evaluation is carried out only in simulation \cite{kim_vehicle_2016}.
	
	The authors of \cite{rabinovich_cobe_2020} propose coded beacons (CoBe) using infrared LEDs for localization and object tracking and obtain convincing results.
	This contribution differs in various aspects from CoBE.
	First, we demonstrate the effectiveness of our system in long range outdoor scenarios, while the performance of CoBE is only investigated indoors. 
	Second, we make use of a bandpass filter, instead of a regular high-pass filter as used by CoBE.
	We found that replacing a bandpass filter with a high-pass filter causes significantly more noise from sun light.
	This makes beacon detection and identification at long ranges more error prone.
	Third, our line code ensures that beacons are never completely turned off.
	We found this to be instrumental for tracking beacons when the vehicle moves at higher speeds.
	Since our beacons are attached to light poles, they cluster closely at long ranges, which increases ambiguity and tracking complexity if they are off for multiple frames.
	Additionally, our approach also significantly differs in the decoding scheme and hardware setup.
	\section{System Design}
	\label{sec:main}
	The underlying idea of this work is to embed active visual beacons into the infrastructure that act as long-term stable, unambiguous, and easily detectable features to facilitate long-term stable vehicle localization in urban environments under changing environment conditions.
	Our privacy-preserving beacons actively communicate a beacon identifier using infrared light.
	Thus, their visibility is not dependent on environmental conditions and external light sources, as would be the case with traffic signs or passive QR codes.
	Furthermore, our system is designed to increase the signal-to-noise ratio, i.e., the visibility of the infrared beacons in contrast to other elements of the environment at the camera sensor.
	This is achieved by restricting the wavelengths of light that reach the camera sensor to a small window around the wavelength emitted by our beacons through the use of a band-pass filter.
	\subsection{Hardware Design}
\begin{figure}
		\centering
		\includegraphics[width=\linewidth]{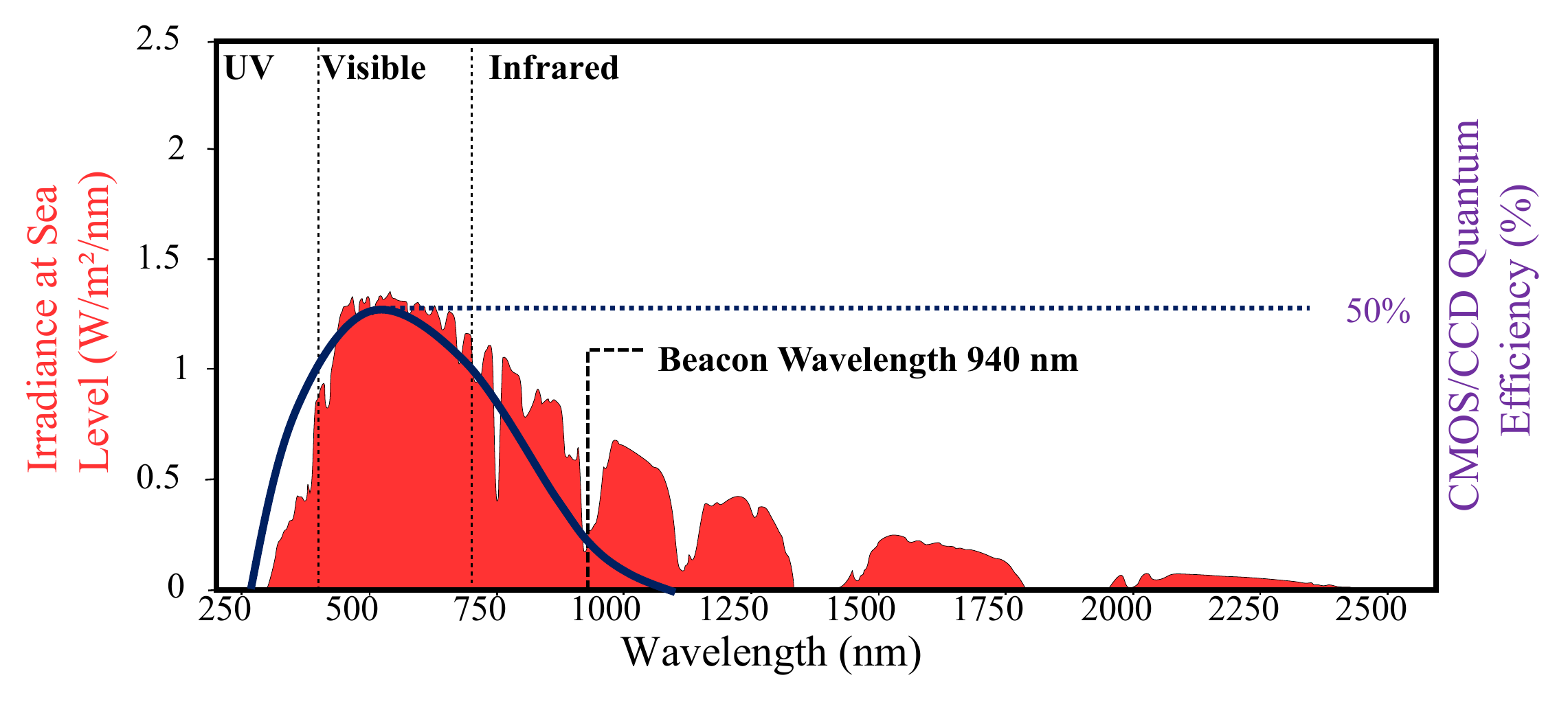}
		\caption{Intensity of sunlight across the wavelength spectrum decreases in the infrared and ultra-violet regions. The selected LED and band-pass combination exploits the intensity drop at \SI{940}{\nano\meter}, where sunlight intensity is low but still detectable for off-the-shelve CMOS/CCD cameras sensors.}
		\label{fig:solar_spectrum}
	\end{figure}

\begin{figure}
		\centering
		\includegraphics{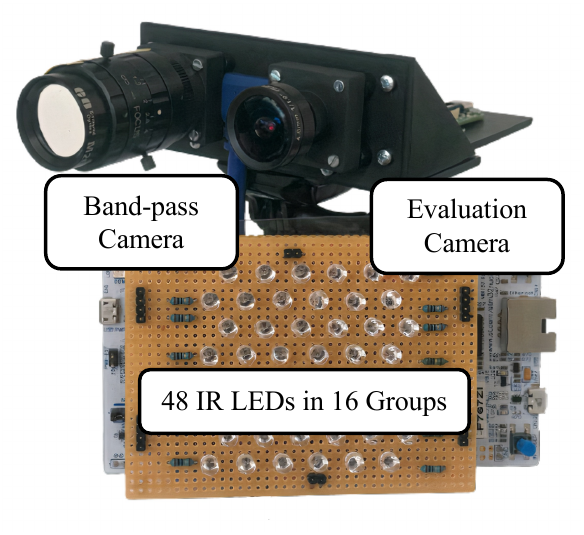}
		\caption{Our prototype beacon and camera system consisting of a camera with optical band-pass and the evaluation camera.}
		\label{fig:hardware_beacon}
	\end{figure}

\begin{figure}[t]
		\centering
		\includegraphics[width=\linewidth]{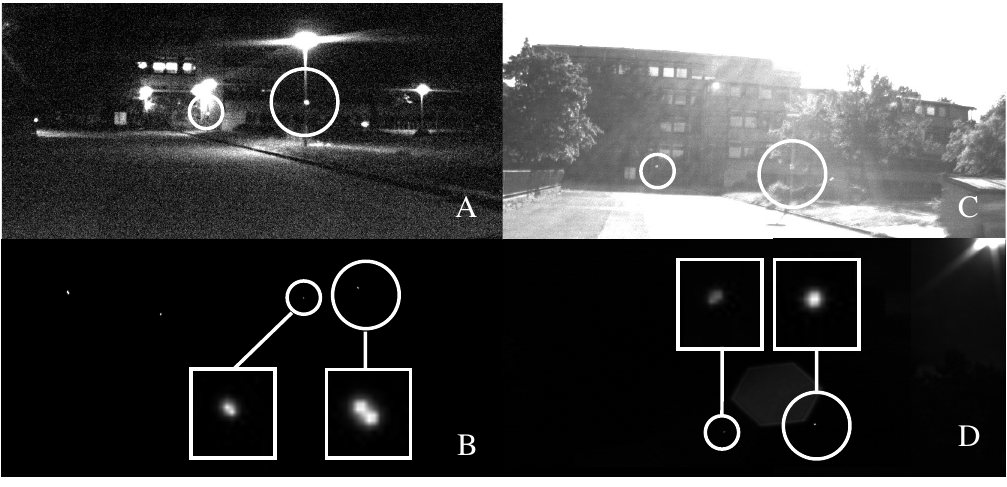}
		\centering
		\caption{Image captured at night (A,B) and against direct sunlight (C,D) with clearly visible beacons in the corresponding band-pass filtered images. Rectangles indicate magnifications of the circled image areas.}
		\label{fig:detections_example}
	\end{figure}

	\textbf{Signal-to-Noise Ratio:} %
	During the daytime, the most dominant cause of interference is sunlight.
	As Fig. \ref{fig:solar_spectrum} shows, the sun as an idealized black-body radiation source emits light across the full wavelength spectrum at varying intensity levels \cite{iqbal_introduction_1983}.
	The sun radiates light with the highest intensity in the human visible spectrum and intensity drops for ultra-violet and infrared light.
	Ideally, the beacon wavelength would be as far as possible in the infrared spectrum to avoid the beacon signal to overlap with natural sunlight.
	Due to physical limitations imposed by the photoelectric effect, quantum efficiency for common, silicon-based camera sensors drops to zero at \SI{1.12}{\electronvolt} photon energy, which corresponds to \SI{1100}{\nano\meter} light wavelength.
	Infrared cameras can overcome this limitation through the use of different sensor principles, but are very expensive and are therefore not considered. 
	However, there are a few characteristic drops in sunlight intensity for specific wavelengths caused by interference with molecules in the atmosphere.
	The drop at \SI{940}{\nano\meter} is particularly strong and is the furthest into the infrared spectrum that is still detectable for regular camera sensors.\\
	\textbf{Beacon and Camera Prototype:} Fig. \ref{fig:hardware_beacon} depicts our proof-of-concept beacons that are made up of 48 LEDs at \SI{940}{\nano\meter} wavelength and measure \SI{6x6}{\centi\meter}.
	In order to allow for investigating different patterns, the LEDs are divided into 16 groups with 3 LEDs each.
	The monochrome camera for beacon detection has a resolution of \SI{1600x1200}{\pixel} and is equipped with a band-pass filter for \SI{950}{\nano\meter} wavelength with a full-width-at-half-maximum value of \SI{50}{\nano\meter}.
	The camera records images at \SI{100}{\hertz}.
	To eliminate remaining \SI{940}{\nano\meter} sunlight, the beacon detection camera uses a low exposure time of \SI{100}{\micro\second}.
	As depicted in Fig. \ref{fig:detections_example}, most of the environment is removed by the filter and the beacons are clearly visible at nighttime and also against direct sunlight during the daytime.
	Most of the surrounding environment is removed from the image, thereby dramatically increasing the signal-to-noise ratio and easing beacon detection.
	\begin{figure}[b]
		\centering
		\includegraphics[width=\columnwidth]{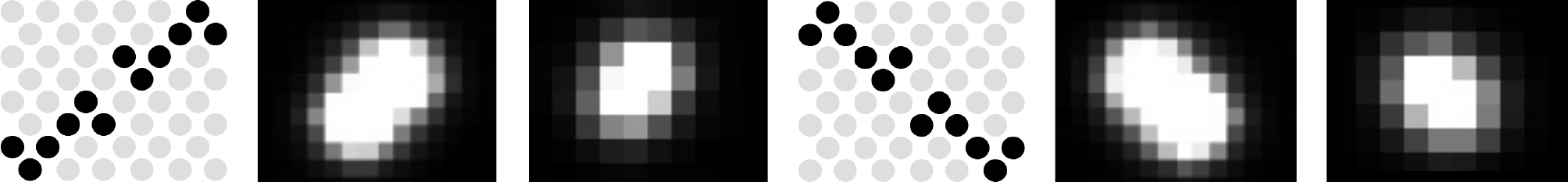}
		\centering
		\caption{A diagonal pattern is used as symbols for data transmission with examples at \SI{25}{\m}  and \SI{50}{\m} as seen by the band-pass camera.}
		\label{fig:symbols}
	\end{figure}
	\subsection{Optical Communication Channel}
	\textbf{Preliminary considerations:} We considered two approaches for designing the communication channel between beacon and camera.
	In a pure spatial encoding approach, each beacon displays a unique pattern for identification (e.g., QR codes). 
	Global identifier uniqueness would not be required as long as it is unique for a sufficiently large local neighborhood of beacons.
	As a result, beacons can potentially be identified from a single camera image instead of having to track and reconstruct the signal over multiple images.
	For this approach, identification of individual LEDs or groups of LEDs is required, which becomes infeasible after a few meters with a \SI{6x6}{\centi\meter} beacon size (see Fig. \ref{fig:detections_example}).
	This approach may be feasible with extreme camera resolutions or significantly larger beacons---options we discarded for practical considerations.
	\\
	\textbf{Linecode:} We choose a mixture of time and spatial encoding, where the identifier is transmitted through the alternation between simple patterns over time.
	An intuitive encoding scheme is turning the beacon on and off in order to transmit 1 or 0.
	Due to the large displacements during the off phase, tracking at a vehicle speed of \SI{30}{\kilo\meter\per\second} has proven difficult in experiments --- especially with multiple beacons.
	Instead, we use the diagonal symbols depicted in Fig. \ref{fig:symbols} to implement a non-return-to-zero line code.
	Depending on diagonal orientation, 1 or 0 is represented.
	The figure shows the symbols at \SI{25}{\meter} and \SI{50}{\meter} as captured by the camera.
	Here it becomes obvious that identifying individual LED groups is not tractable for a pure spatial domain encoding.
	The chosen approach has the advantage that the beacons are always active, which eases tracking and subsequently signal reconstruction.\\
	\textbf{Beacon identifier:} A 12-Bit-Code is assigned to every beacon, which is transmitted in an endless loop using the two previously introduced symbols, see Fig. \ref{fig:symbols}.
	We do not use a synchronization symbol.
	Ambiguities arise if codes are chosen unconstrained, e.g. \texttt{00110011} and \texttt{11001100} generate the same signal if repeated indefinitely.
	Therefore, we use prefix-free codes that are generated through ambiguity testing of all cyclic shifts for all  $2^{12}$ possible identifiers.
	The camera records an image every \SI{10}{\ms} while the beacon displays each bit for \SI{70}{\ms}, which allows for oversampling and more robust signal reconstruction.
	Transmission of the full 12-bit beacon identifier takes \SI{0.84}{\s}.
	
	\subsection{Image Processing}
	We will now describe the image processing pipeline used for beacon detection and identification.
	The processing pipeline generates image patches that potentially contain beacons.
	Candidates are tracked across multiple images and the signal is reconstructed.\\
	\textbf{Beacon Detection:}
	The beacon detection stage first operates on binarized images for filtering purposes, but uses grayscale images for subsequent stages. %
	We first generate a set of proposal image patches that contain potential beacons and filter the proposals based on reference image comparison.
	First, the 8-bit greyscale image $I_G$ is converted to a binary image $I_B$ by setting pixel intensities lower than $5$ to zero, and pixel values above $5$ to $1$.
	Next, a set of contiguous areas are extracted from the camera image using morphological operations as beacon proposals $P = \{(x_1,y_1,w_1,h_1),...,(x_n,y_n,w_n,h_n)\}$.
	Each area is described through its bounding box width, height and location in the camera image.
	
	As beacons have a fixed size, the maximum and minimum area occupied in the camera image depends only on the distance to the camera.
	The upper and lower bounds can be determined analytically using the pinhole camera model.
	In theory, a beacon at \SI{60}{\metre} occupies only \SI{2x2}{\pixel} according to the pinhole model.
	However, due to the fact that neighboring pixels are also illuminated, the beacon orientation is still distinguishable at well over \SI{100}{\metre} (see Section \ref{sec:evaluation}).
	The proposals $P$ are filtered based on these limits such that both large-scale artifacts and individual white pixels are excluded.
	The set of proposals is refined to $P_R$ based on proposal area: $P_R = \{ (x,y,w,h) \in P \mid wh \in [3,400]\}$.
	For distant beacons, individual pixels of low intensity have been removed by thresholding, but the decoder is still able to determine the orientation correctly in the grayscale image $I_G$.
	\\
	The final detection step uses shape matching to compare the proposals with a reference image $I_R$.
	To this end, we compute Hu-moments for each beacon $b \in P_R$ in the original grayscale image $I_G$, which are translation, rotation and scale-invariant \cite{1057692}.
	Image moments of order $i+j$ are defined as follows:
	\begin{equation}
		\mu_{ij} = \sum_x \sum_y (x - \bar x)^i (y - \bar y)^j I_G(x,y)
	\end{equation}
	with $\bar x$ and $\bar y$ being the centroids in each dimension:
	\begin{equation}
		\bar x = \frac{\sum_x \sum_y x I_G(x,y)}{\sum_x \sum_y I_G(x,y)} %
	\end{equation}
	\begin{equation}
		\bar y = \frac{\sum_x \sum_y y I_G(x,y)}{\sum_x \sum_y I_G(x,y)} %
	\end{equation}
	Scale invariant moments are then obtained through normalization as:
	\begin{equation}
		\eta_{ij} = \frac{\mu_{ij}}{\mu_{00}^{(\frac{i+j}{2} + 1)}}
	\end{equation}
	The seven rotational invariant Hu-moments $c_1,\ldots,c_7$ can then be defined using $\eta_{ij}$.
	For example $c_1 = \eta_{20} + \eta_{02}$ and  $c_2 = (\eta_{20} - \eta_{02})^2 + 4\eta_{11}^2$.
	A complete list of all moment definitions is provided in \cite{1057692}.
	The distance between $I_R$ and a proposal patch $I_P$ extracted from $I_G$ is then computed on the basis of respective Hu-moments:
	\begin{equation}
		\begin{split}
			D(I_P,I_R)=\sum_{i=1}^7\biggl| \frac{1}{\sgn(c^P_i) \log c^P_i}-\frac{1}{\sgn(c^R_i) \log c^R_i}\biggl|
		\end{split}
	\end{equation}
	Proposals for which $D(I_P,I_R) < 0.2$ holds are regarded as detections $d \in D$ and are passed to the tracker.
	\\
	\textbf{Tracking:}
	Due to the high frame rate of \SI{100}{\hertz}, we assume only small shifts of the beacon position between consecutive images.
	Input to this stage are tracked beacons from previous images denoted as tracks $T=\{t_1,\ldots,t_l\}$ and detections for the current camera image $D=\{d_1,\ldots,d_k\}$.
	We compute a distance matrix for each detection and track combination based on the euclidean distance of bounding box centroids as follows:
		We follow a greedy approach and associate tracks and detections starting with the smallest distance up to a maximum distance threshold.
		A new track is created for all detections that can not be matched to any existing track, either due to exceeding the maximum distance or because all other tracks have been matched.
		Tracks are discarded if they are not matched to a detection for more than 30 frames.\\
		\textbf{Signal Reconstruction:}
		The decoder attempts to recover the identifier for each track. 
		For this, the orientation of the symbol is determined based on the central image moments:
		\begin{equation}
			\label{orientation_estimate}
			\theta = \frac{1}{2} \arctan{\bigg( \frac{2 \mu_{11}}{\mu_{20} - \mu_{02}} \bigg)}
		\end{equation}
		Orientations are mapped to $1$ if $\theta > 0$ and to $0$ otherwise.
		The decoder is designed to make use of the oversampled symbol display, i.e., \SI{10}{\milli\second} time between camera images and \SI{70}{\milli\second} display time for a single bit.
		For each track, a sliding window sums up the last seven estimated bit values (c.f.  Eq. \ref{orientation_estimate}).
		\begin{figure}
			\centering
			\includegraphics[width=\linewidth]{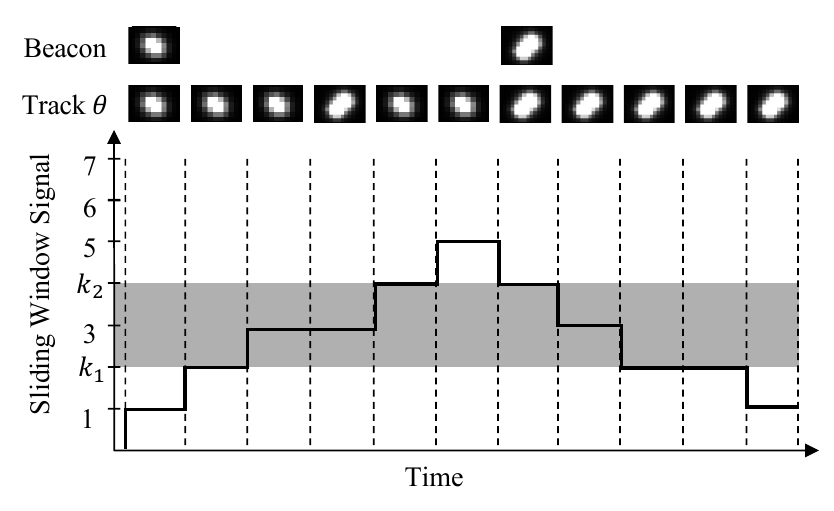}
			\centering
			\caption{Signal reconstruction uses a sliding window summation over the previous orientation estimates. A Schmitt trigger evaluates the sliding window signal.}
			\label{fig:decoder}
		\end{figure}
		As depicted in Figure \ref{fig:decoder}, the sliding window produces a signal $n$ that is input to a Schmitt trigger, with threshold values $k_1 = 2$ and $k_2 = 4$.
		If the sum exceeds $k_2$, the beacon signal is interpreted as \texttt{1}, and if it falls below $k_1$ as \texttt{0}.
		Whenever a signal level change has been detected, a single bit is appended to the decoded identifier of the track corresponding to the Schmitt trigger state and the time of signal change is stored.
		The decoder also handles repetitions of the same symbol, as the time to sample the Schmitt trigger is derived from the known beacon frequency.
		Each track $t_i$ generates a bit sequence that can be matched to a known list of beacon identifiers.
		\section{Experiments \& Results}
\begin{figure}[b]
			\centering
			\includegraphics{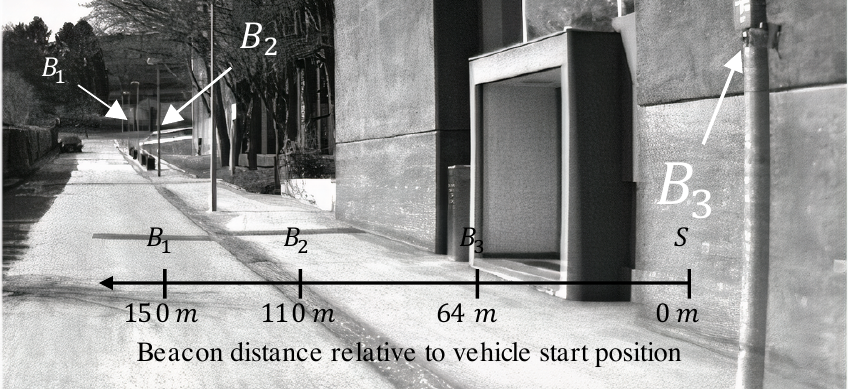}
			\centering
			\caption{Our experiment consists of beacons $B_1, B_2$ and $B_3$ mounted \SI{40}{\meter} apart on average. The vehicle starts \SI{150}{\meter} away from $B_1$.}
			\label{fig:setup}
		\end{figure}

		The following experiments are conducted on a straight road.
		As displayed in Fig. \ref{fig:setup}, we have mounted beacons $B_1,B_2$ and $B_3$ on light poles that are on average \SI{40}{\meter} apart.
		Fig. \ref{fig:setup} shows the distance of each beacon to the starting point $S$, with $B_1$ being the furthest away at \SI{150}{\meter}.
		A video that demonstrates the experiments is available in \cite{InvestigatingOutdoorRecognition}.
		\label{sec:evaluation}
		\begin{table}[t]
			\caption{Performance at various distances during standstill}
			\label{tab:standstill}
			\begin{center}
				\begin{tabular}{@{}llllll@{}}
					\toprule
					Metric &  $\SI{40}{\meter}$ & $\SI{60}{\meter}$ & $\SI{80}{\meter}$ & $\SI{100}{\meter}$ & $\SI{120}{\meter}$ \\ 
					\midrule 
					Frames 		      &  2264 	&  2279 & 2277 &  2278 &  2276\\ 
					Detections      & 2264 	   &  2279 & 2277 & 2278 &  2269 \\ 
					Tracks 		  	   & 1    	     & 1         & 1        & 1        & 1 \\ 
					Bits Decoded  & 263       & 265    &  263   & 261   &  271 \\ 
					Error Bits 			&  0 		 &  0 		 & 0	    & 0 	   & 104 \\ 
					\bottomrule 
				\end{tabular} 
			\end{center}
		\end{table}
		
		\subsection{Standstill Experiment}
		\begin{table*}
			\caption{Evaluation at daytime $D_i$ and nighttime $N_i$. The vehicle position upon first and last correct identifier recognition is reported. Beacons are detected over a distance of more than \SI{100}{\m}.}
			\label{tab:moving}
			\begin{center}
				\begin{tabular}{@{}rrrrrrcrrrrr@{}}
					\toprule
					&  $D_1$ &  $D_2$ & $D_3$ & $D_4$ & Avg. & & $N_1$ & $N_2$  &  $N_3$  & $N_4$ & Avg.  \\ 
					\midrule 
					Frames  & 2940 & 2771 & 2723 & 2626 & 2765.0  &  & 2739 & 2678 & 2643 & 2739 & 2699.75 \\ 
					Detections & 5247 & 5026 & 5097 & 4935 & 5076.25  &  & 4958 & 4895 & 4667 & 4736 & 4814.0 \\ 
					Tracks  & 10 & 9 & 12 & 11 & 10.5  &  & 5 & 5 & 8 & 7 & 6.25 \\ 
					
					\midrule
					$B_1$ Tracks  & 1 & 1 & 1 & 1 & 1.0  & & 1 & 2 & 1 & 1 & 1.25 \\
					$B_1$ First Recognition  & \SI{50.9}{\m}  & \SI{51.8}{\m}  & \SI{50.2}{\m}  & \SI{51.4}{\m}  & \SI{51.07}{\m}   & & \SI{14.5}{\m}  & \SI{18.5}{\m}  & \SI{64.6}{\m}  & \SI{45.0}{\m}  & \SI{35.65}{\m}  \\
					$B_1$ Last Recognition  & \SI{145.0}{\m}  & \SI{145.1}{\m}  & \SI{149.9}{\m}  & \SI{145.3}{\m} & \SI{146.32}{\m}   & & \SI{148.2}{\m}  & \SI{148.0}{\m}  & \SI{148.0}{\m}  & \SI{148.5}{\m} & \SI{148.18}{\m}  \\
					$B_1$ Bits Decoded  & 213  & 217  & 231  & 194  & 213.75  &  & 279  & 263  & 172  & 209  & 230.75 \\ 
					$B_1$ Error Bits  & 33  & 49  & 75  & 17  & 43.5  &  & 51  & 71  & 11  & 29  & 40.5 \\ 
					
					\midrule
					$B_2$ Tracks  & 1 & 1 & 1 & 1 & 1.0  & & 1 & 1 & 1 & 1 & 1.0 \\
					$B_2$ First Recognition  & \SI{4.0}{\m}  & \SI{5.5}{\m}  & \SI{6.3}{\m}  & \SI{7.7}{\m}  & \SI{5.88}{\m}   & & \SI{6.8}{\m}  & \SI{6.5}{\m}  & \SI{8.2}{\m}  & \SI{10.0}{\m}  & \SI{7.88}{\m}  \\
					$B_2$ Last Recognition  & \SI{105.6}{\m}  & \SI{105.5}{\m}  & \SI{105.2}{\m}  & \SI{105.0}{\m} & \SI{105.33}{\m}   & & \SI{105.9}{\m}  & \SI{105.9}{\m}  & \SI{105.5}{\m}  & \SI{105.4}{\m} & \SI{105.68}{\m}  \\
					$B_2$ Bits Decoded  & 234  & 213  & 216  & 211  & 218.5  &  & 223  & 221  & 217  & 221  & 220.5 \\ 
					$B_2$ Error Bits  & 6  & 7  & 18  & 10  & 10.25  &  & 10  & 11  & 14  & 29  & 16.0 \\ 
					
					\midrule
					$B_3$ Tracks  & 1 & 1 & 1 & 1 & 1.0  & & 1 & 1 & 1 & 1 & 1.0 \\
					$B_3$ First Recognition  & \SI{4.8}{\m}  & \SI{7.2}{\m}  & \SI{8.4}{\m}  & \SI{5.4}{\m}  & \SI{6.45}{\m}   & & \SI{7.1}{\m}  & \SI{5.7}{\m}  & \SI{8.6}{\m}  & \SI{7.1}{\m}  & \SI{7.12}{\m}  \\
					$B_3$ Last Recognition  & \SI{62.1}{\m}  & \SI{62.4}{\m}  & \SI{61.7}{\m}  & \SI{62.1}{\m} & \SI{62.07}{\m}   & & \SI{63.5}{\m}  & \SI{63.8}{\m}  & \SI{63.0}{\m}  & \SI{63.0}{\m} & \SI{63.33}{\m}  \\
					$B_3$ Bits Decoded  & 151  & 132  & 131  & 131  & 136.25  &  & 141  & 142  & 133  & 141  & 139.25 \\ 
					$B_3$ Error Bits  & 0  & 5  & 0  & 20  & 6.25  &  & 12  & 5  & 4  & 12  & 8.25 \\
					\bottomrule 
				\end{tabular} 
			\end{center}
		\end{table*}
		This experiment intends to determine the system performance when the vehicle is not moving.
		For this purpose, only beacon $B_1$ is activated and we have parked our test vehicle $\SI{40}{\meter}$, $\SI{60}{\meter}$, $\SI{80}{\meter}$, $\SI{100}{\meter}$, and $\SI{120}{\meter}$ away from $B_1$.
		We investigate the ability of the system to correctly decode the identifier transmitted by the beacon.
		This experiment evaluates the end-to-end performance of the system since a correctly decoded beacon identifier requires all stages of the processing pipeline to function correctly.
		All recordings for this experiment have been recorded during daytime.%
		
		The outcome of this experiment is summarized in Table~\ref{tab:standstill}.
		Although the experiment was conducted in daylight, no detections and thus tracks were generated other than the one corresponding to the beacon for all distances.
		The total number of bits decoded corresponds to the length of the bit-string that is generated by the signal reconstruction stage.
		
		We count all occurrences of the known beacon identifier in the decoded bit-string. 
		The first and last bits are also considered correct if they are a part of the true identifier. 
		All remaining bits are considered error bits.
		For example, if we decode \texttt{11001000010011001010001001100100001}, then \texttt{110010}\textbf{\texttt{X}}\texttt{1}\textbf{\texttt{X}}\texttt{0001} results after replacing all occurrences of $id(B_1) = 000100110010$ with \texttt{X}.
		Since \texttt{110010} and \texttt{0001} are part of $id(B_1)$, we count the number of correctly decoded bits as 34 and the number of error bits as 1.
		
		The results show that the system correctly derives the beacon identifier up to a distance of $\SI{100}{\meter}$ error-free.
		We see that the system is also able to detect and track the beacon at a distance of $\SI{120}{\meter}$.
		The fact that there are less detections than frames indicates that the detector fails to generate beacon candidates, with a negative impact on the decoded bit sequence.
		Although we were still able to find $id(B_1)$ in the decoded bit string multiple times at $\SI{120}{\meter}$, we see that errors start to occur.
		For practical applications, this might still be sufficient.
		While the position of $B_1$ in the camera image is crucial for ego-pose estimation, querying the geographical location of the beacon using the identifier only has to be done once.
		\subsection{Driving Experiment}
		
		This experiment determines the ability of the system to detect and identify the beacons while driving. 
		The vehicle accelerates from a standstill at position $S$ to $\SI{8.3}{\meter\per\second}$ and drives towards the end of the road until it passes $B_1$.
		All three beacons are active with $id(B_2) = 010100100110$ and $id(B_3) = 000101010100$.
		We have conducted eight runs of this experiment, four during daytime and four at nighttime.  
		
		The results for this experiment are shown in Table \ref{tab:moving}.
		We report the total number of frames, detections, tracks as well as total bits decoded for each recording.
		The number of detections per run appear large, but it is less than three beacons per frame.
		This is expected, as $B_2$ and $B_3$ leave the field of view and $B_1$ is picked up only after \SI{50}{\meter}.
		Furthermore, we report the number of tracks for each beacon where the generated decoder bit sequence contains the correct beacon identifier.
		In each run, only one track generates the correct identifier per beacon, with an exception of run $N_2$.
		In $N_2$, beacon $B_1$ is recognized very early.
		This early generated track is discarded as the detector fails to find $B_1$.
		Later on, $B_1$ is picked up at the same distance as in the other experiments.
		
		We also show the vehicle position (cf. Fig. \ref{fig:setup}) of the first and last recognition of the full beacon identifier.
		$B_3$ and $B_2$ are identified within the \SI{100}{\meter} recognition range observed during the standstill experiment.
		The detection distance is non-zero, as the vehicle is in motion and the transmission of the 12-bit identifier takes \SI{0.84}{\second}.
		At \SI{8.3}{\meter\per\second}, the vehicle drives \SI{6.9}{\meter} during a complete identifier transmission.
		The observed numbers are slightly lower than in the standstill experiment.
		
		Table \ref{tab:moving} also shows the correct and incorrect bits in the decoded bit string as defined in the standstill  experiment.
		Here, we can observe a higher number of error bits.
		We found that in many cases the errors occur at the end of track lifetime.
		As beacons leave the field of view, bit are mistakenly appended as the decoder assumes symbol repetition.
		
		The performance of the system does not differ significantly during nighttime or daytime, which becomes clear when comparing decoding errors.
		This might be attributed to normalization of the camera image achieved by the band-pass filter.
		Due to the absence of light sources at \SI{940}{\nano\meter} wavelength, the number of detections and thus the number of tracks is slightly lower at nighttime.
		It is noteworthy that $id(B_1)$ is recognized on average \SI{15,4}{\meter} earlier at nighttime in comparison to daytime.
		The greater recognition distance might be attributed to less noisy detections and therefore more robust template matching in the detector.

		\addtolength{\textheight}{-5.5cm}
		\section{CONCLUSION \& FUTURE WORK}
		\label{sec:conclusion}
		We have presented an approach for infrastructure-based localization using infrared beacons and a band-pass filtered camera system.
		The system setup and underlying design considerations have been elaborated.
		We have described our image processing pipeline that uses traditional computer vision techniques.
		Our experiments demonstrate the ability of the system to detect, track and identify our beacons independent of light conditions at long ranges.
		As our detection range is greater than the typical distance between light poles in urban areas, the obtrusive and costly construction of new infrastructure is not required.
		The use of infrared light increases signal-to-noise ratio and is invisible to the human eye.
		Our contribution can be regarded as a proof-of-concept but more large-scale investigations have to be carried out.
		These experiments will also investigate the final positioning performance achieved.
		Additionally, we plan to investigate alternative beacon patterns, as our beacon prototype allows to display up to $65{,}535$ different symbols.
		\bibliography{references.bib,zotero_export.bib}{}
		\bibliographystyle{ieeetr}
	\end{document}